\documentclass[11pt]{article}
\usepackage{arabtex}
\usepackage{utf8}
\usepackage{coling2018}
\usepackage{times}
\usepackage{url}
\usepackage{graphicx} % use \rotatebox{90}{rota}
\usepackage{pgfplots}

\usetikzlibrary{shapes,backgrounds}
\pgfplotsset{compat=1.14}

\usepackage[stable]{footmisc}
\date{}

\setcode{utf8}
\begin{document}

\title{\textbf{Kurdish (Sorani) Speech to Text: \\Presenting an Experimental Dataset}}

\author{
\begin{tabular}[t]{c c}
{Akam Qader} & Hossein Hassani\\
\textnormal{University of Kurdistan Hewl\^er} & \textnormal{University of Kurdistan Hewl\^er}\\
\textnormal{Kurdistan Region - Iraq} & 
\textnormal{Kurdistan Region - Iraq}\\
{\tt } & {\tt hosseinh@ukh.edu.krd}
\end{tabular}
}

\maketitle

\vskip 1.25cm

\begin{abstract}
	We present an experimental dataset, Basic Dataset for Sorani Kurdish Automatic Speech Recognition (BD-4SK-ASR), which we used in the first attempt in developing an automatic speech recognition for Sorani Kurdish. The objective of the project was to develop a system that automatically could recognize simple sentences based on the vocabulary which is used in grades one to three of the primary schools in the Kurdistan Region of Iraq. We used CMUSphinx as our experimental environment. We developed a dataset to train the system. The dataset is publicly available for non-commercial use under the \texttt{CC BY-NC-SA 4.0} license\footnote{\url{https://creativecommons.org/licenses/by-nc-sa/4.0/}}.
\end{abstract}

\section{Introduction}
\label{intro}
Kurdish language processing requires endeavor by interested researchers and scholars to overcome with a large gap which it has regarding the resource scarcity. The areas that need attention and the efforts required have been addressed in \cite{hassani2018blark}.

\par The Kurdish speech recognition is an area which has not been studied so far. We were not able to retrieve any resources in the literature regarding this subject.
 
\par In this paper, we present a dataset based on CMUShpinx~\cite{cmushinx4} for Sorani Kurdish. We call it a Dataset for Sorani Kurdish Automatic Speech Recognition (BD-4SK-ASR). Although other technologies are emerging, CMUShpinx could still be used for experimental studies.

\par The rest of this paper is organized as follows. Section \ref{relatedwork} reviews the related work. Section \ref{dataset} presents different parts of the dataset, such as the dictionary, phoneset, transcriptions, corpus, and language model. Finally, Section \ref{conclusion} concludes the paper and suggests some areas for future work.

\section{Related work}
\label{relatedwork}

The work on Automatic Speech Recognition (ASR) has a long history, but we could not retrieve any literature on Kurdish ASR at the time of compiling this article. However, the literature on ASR for different languages is resourceful. Also, researchers have widely used CMUSphinx for ASR though other technologies have been emerging in recent years \cite{cmushinx4}.

\par We decided to use CMUSphinx because we found it a proper and well-established environment to start Kurdish ASR.

\section{The BD-4SK-ASR Dataset}
\label{dataset}
To develop the dataset, we extracted 200 sentences from Sorani Kurdish books of grades one to three of the primary school in the Kurdistan Region of Iraq. We randomly created 2000 sentences from the extracted sentences. 

\par In the following sections, we present the available items in the dataset. The dataset ia available on \url{https://github.com/KurdishBLARK/BD-4SK-ASR}.

\subsection{Phoeset}
The phoneset includes 34 phones for Sorani Kurdish. A sample of the file content is given below.

\begin{center}
	\texttt{R\\
		RR\\
		S\\
		SIL\\
		SH\\
		T\\
		V\\
		W\\
		WW
	}
\end{center}

\begin{figure}
	\centering
	\includegraphics[width=0.6\linewidth]{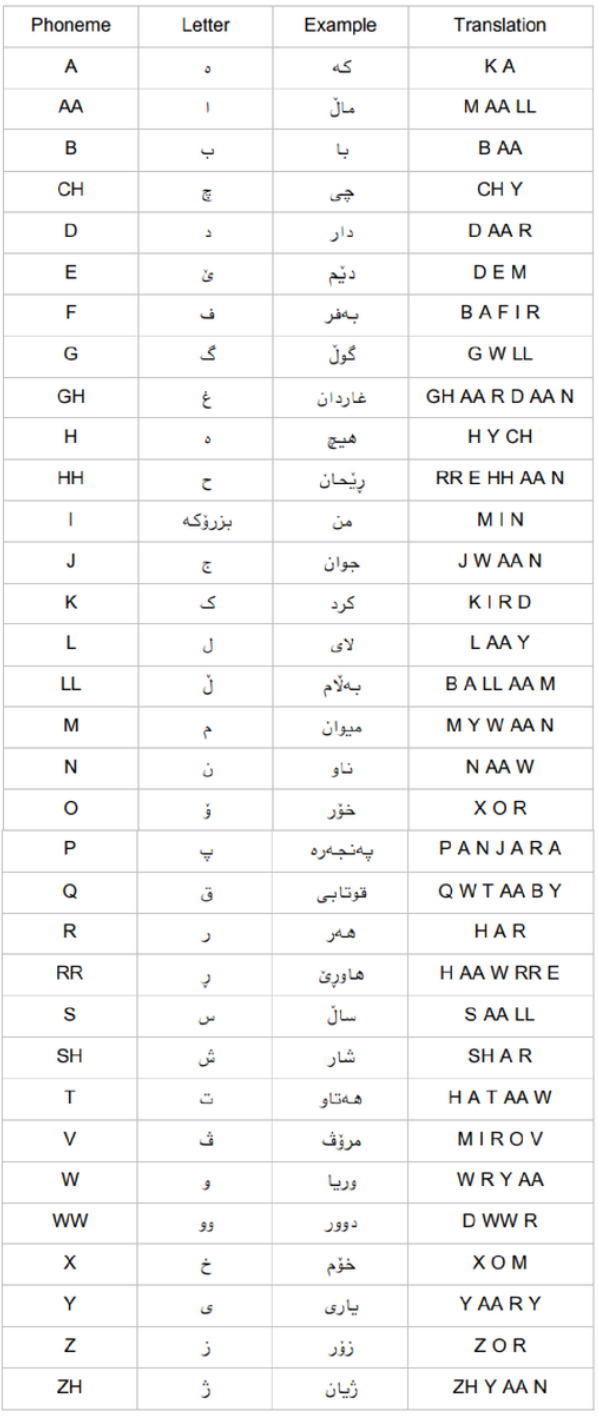}
	\caption{The Sorani sounds along with their phoneme representation.}
	\label{fig:phone}
\end{figure}

Figure \ref{fig:phone} shows the Sorani letters in Persian-Arabic script, the suggested phoneme (capital English letters), and an example of the transformation of words in the developed corpus.

\subsection{Filler phones}
The filler phone file usually contains fillers in spoken sentences. In our basic sentences, we have only considered silence. Therefore it only includes three lines to indicate the possible pauses at the beginning and end of the sentences and also after each word.

\subsection{The File IDs}
This file includes the list of files in which the narrated sentences have been recorded. The recorded files are in \textit{wav} formats. However, in the file IDs, the extension is omitted. A sample of the file content is given below. The \textit{test} directory is the directory in which the files are located.

\begin{center}
	\texttt{test/T1-1-50-01\\
		test/T1-1-50-02\\
		test/T1-1-50-03\\
		test/T1-1-50-04\\
		test/T1-1-50-05\\
		test/T1-1-50-06}
\end{center}

\subsection{The Transcription}
\label{subsec:trans}
This file contains the transcription of each sentence based on the phoneset along with the file ID in which the equivalent narration has been saved. The following is a sample of the content of the file.

\begin{flushleft}
	
	\begin{quotation}
		\texttt{<s> BYR RRAAMAAN DAARISTAANA AMAANAY </s> (T1-1-50-18)\\
		<s> DWWRA HAWLER CHIRAAYA SARDAAN NABWW </s> (T1-1-50-19)\\
		<s> SAALL DYWAAR QWTAABXAANA NACHIN </s> (T1-1-50-20)\\
		<s> XWENDIN ANDAAMAANY GASHA </s> (T1-1-50-21)\\
		<s> NAMAAM WRYAA KIRD PSHWWDAA </s> (T1-1-50-22)\\
		<s> DARCHWWY DAKAN DAKAWET </s> (T1-1-50-23)\\
		<s> CHAND BIRAAT MAQAST </s> (T1-1-50-24)\\
		<s> BAAXCHAKAY DAAYK DARCHWWY </s> (T1-1-50-25)\\
		<s> RROZH JWAAN DAKAWET ZYAANYAAN </s> (T1-1-50-26)}\\
	\end{quotation}

\end{flushleft}

\subsection{The Corpus}
The corpus includes 2000 sentences. Theses sentence are random renderings of 200 sentences, which we have taken from Sorani Kurdish books of the grades one to three of the primary school in the Kurdistan Region of Iraq. The reason that we have taken only 200 sentences is to have a smaller dictionary and also to increase the repetition of each word in the narrated speech. We transformed the corpus sentences, which are in Persian-Arabic script, into the format which complies with the suggested phones for the related Sorani letters (see Section~\ref{subsec:trans}).  

\subsection{The Narration Files}
Two thousand narration files were created. We used Audacity\footnote{https://www.audacityteam.org/} to record the narrations. We used a normal laptop in a quiet room and minimized the background noise. However, we could not manage to avoid the noise of the fan of the laptop. A single speaker narrated the 2000 sentences,  which took several days. We set the Audacity software to have a sampling rate of 16, 16-bit bit rate, and a mono (single) channel. The noise reduction db was set to 6, the sensitivity to 4.00, and the frequency smoothing to 0.

\subsection{The Language Model}
We created the language from the transcriptions. The model was created using CMUSphinx in which (fixed) discount mass is 0.5, and backoffs are computed using the ratio method. The model includes 283 unigrams, 5337 bigrams, and 6935 trigrams.

\section{Conclusion}
\label{conclusion}
We presented a dataset, BD-4SK-ASR, that could be used in training and developing an acoustic model for Automatic Speech Recognition in CMUSphinx environment for Sorani Kurdish. The Kurdish books of grades one to three of primary schools in the Kurdistan Region of Iraq were used to extract 200 sample sentences. The dataset includes the dictionary, the phoneset, the transcriptions of the corpus sentences using the suggested phones, the recorded narrations of the sentences, and the acoustic model. The dataset could be used to start experiments on Sorani Kurdish ASR.

\par As it was mentioned before, research and development on Kurdish ASR require a huge amount of effort. A variety of areas must be explored, and various resources must be collected and developed. The multi-dialect characteristic of Kurdish makes these tasks rather demanding. To participate in these efforts, we are interested in the expansion of Kurdish ASR by developing a larger dataset based on larger Sorani corpora, working on the other Kurdish dialects, and using new environments for ASR such as Kaldi\footnote{https://kaldi-asr.org/}.  

\bibliographystyle{lrec}
\bibliography{kstt2018}

\begin{thebibliography}{}

\bibitem[\protect\citename{{CMUSphinx}}2019]{cmushinx4}
{CMUSphinx}.
\newblock (2019).
\newblock {Open Source Speech Recognition Toolkit}.
\newblock \url {https://cmusphinx.github.io/}.
\newblock [Accessed on: November 28, 2019].

\bibitem[\protect\citename{Hassani}2018]{hassani2018blark}
Hassani, H.
\newblock (2018).
\newblock {BLARK for multi-dialect languages: towards the Kurdish BLARK}.
\newblock {\em Language Resources and Evaluation}, 52:625--644.

\end{thebibliography}

\end{document}